\documentclass[letterpaper]{article} 
\usepackage{aaai25}  
\usepackage{times}  
\usepackage{helvet}  
\usepackage{courier}  
\usepackage[hyphens]{url}  
\usepackage{graphicx} 
\usepackage{natbib}  
\usepackage{caption} 
\usepackage{algorithm}
\usepackage{algorithmic}

\usepackage{newfloat}
\usepackage{listings}
\DeclareCaptionStyle{ruled}{labelfont=normalfont,labelsep=colon,strut=off} 
\floatstyle{ruled}
\newfloat{listing}{tb}{lst}{}
\floatname{listing}{Listing}
\usepackage{subfigure}
\usepackage{colortbl}
\usepackage[table]{xcolor}
\usepackage{pdfpages}
\usepackage[utf8]{inputenc}
\usepackage{cite}

\usepackage{xcolor}
\usepackage{eqnarray}
\usepackage{algorithm}
\usepackage{algorithmic}
\usepackage{amstext,amsmath,amsfonts,amssymb,amsthm,tikz,tikz-inet,pgf,pgfplots}
\usepackage[capitalize]{cleveref}
\usepackage{float}
\usepackage{bbold}
\usepackage{answers}
\pgfplotsset{compat=1.14}
\usepgfplotslibrary{statistics}
\usepackage{url}
\usepackage{multirow}
\usetikzlibrary{automata,chains}
\usepackage{svg}

\usepackage{tikz}
\usepackage{amsmath}
\usepackage{amssymb}
\usepackage{array}
\usepackage{rotating}

\crefname{definition}{Definition}{Definitions}
\crefname{proposition}{Proposition}{Propositions}
\crefname{propriete}{Propriété}{Propriétés}
\crefname{hypothese}{Hypothesis}{Hypotheses}
\crefname{algorithm}{Algorithm}{Algorithms}
\crefname{lemme}{Lemma}{Lemmas}
\crefname{notation}{Notation}{Notations}
\crefname{table}{Table}{Tables}
\crefname{assertion}{assertion}{assertions}

\title{Search Strategy Generation for Branch and Bound Using Genetic Programming}
\author{Gwen Maudet\textsuperscript{\rm 1}, Grégoire Danoy\textsuperscript{\rm 2}}
\affiliations {\textsuperscript{\rm 1}SnT, University of Luxembourg, Esch-sur-Alzette, Luxembourg\\
\textsuperscript{\rm 2}FSTM/DCS, SnT, University of Luxembourg, Esch-sur-Alzette, Luxembourg\\
\{gwen.maudet,gregoire.danoy\}@uni.lu}
\date{\today}
\begin{document}

\maketitle

\begin{abstract}
Branch-and-Bound (B\&B) is an exact method in integer programming that recursively divides the search space into a tree. During the resolution process, determining the next subproblem to explore within the tree—known as the search strategy—is crucial. 
Hand-crafted heuristics are commonly used, but none are effective over all problem classes. 
Recent approaches utilizing neural networks claim to make more intelligent decisions but are computationally expensive.
In this paper, we introduce GP2S (Genetic Programming for Search Strategy), a novel machine learning approach that automatically generates a B\&B search strategy heuristic, aiming to make intelligent decisions while being computationally lightweight. 
We define a policy as a function that evaluates the quality of a B\&B node by combining features from the node and the problem; the search strategy policy is then defined by a best-first search based on this node ranking. 
The policy space is explored using a genetic programming algorithm, and the policy that achieves the best performance on a training set is selected.
We compare our approach with the standard method of the SCIP solver, a recent graph neural network-based method, and handcrafted heuristics. 
Our first evaluation includes three types of primal hard problems, tested on instances similar to the training set and on larger instances. 
Our method is at most 2\% slower than the best baseline and consistently outperforms SCIP, achieving an average speedup of 11.3\%. Additionally, GP2S is tested on the MIPLIB 2017 dataset, generating multiple heuristics from different subsets of instances. It exceeds SCIP’s average performance in 7 out of 10 cases across 15 times more instances and under a time limit 15 times longer, with some GP2S methods leading on most experiments in terms of the number of feasible solutions or optimality gap.

\end{abstract}
\section{Introduction}

Mixed-integer linear programming (MILP) is crucial for optimizing various sectors like transportation~\cite{archetti_optimization_2022,yang_survey_2016}, production~\cite{tiwari_survey_2015}, and e-health services~\cite{rais_operations_2011}, by minimizing or maximizing an objective function under specific constraints. While exact algorithms ensure optimal solutions and quantify proximity to the optimum, they are computationally intensive. Branch-and-Bound (B\&B), the standard algorithm for solving discrete optimization problems, utilizes a "divide and conquer" approach, partitioning the solution space and constructing a tree where simpler subproblems are solved at deeper levels.

B\&B has key components that guide decision-making: \emph{cutting} (adding constraints), \emph{branching} (choosing variables to split the search space), and \emph{searching} (determining exploration order within the B\&B tree). The solver aims to quickly find high-quality solutions and prove optimality. However, decision-making in B\&B remains an open challenge, with no universal method effective across diverse instances~\cite{zarpellon_parameterizing_2021,parsonson_reinforcement_2023}. The choice of search strategy (SS)~\cite{ibaraki_theoretical_1976,morrison_branch-and-bound_2016}—the focus of this paper—impacts solution updates and pruning efficiency, significantly affecting overall execution time. Current methods rely on manually crafted heuristics, which lack consistency across MILP instances~\cite{linderoth_computational_1999}. Recent research has explored ML-based SS policies, often using neural networks to mimic oracle decisions~\cite{he_learning_2014,song_learning_2018,yilmaz_study_2021,labassi_learning_2022}, but these approaches are constrained by the oracle's limitations and the computational overhead of neural networks.

This paper introduces \emph{GP2S} (Genetic Programming for Search Strategy), a method designed to automatically generate a SS policy using genetic programming (GP), aiming to be both lightweight and efficient. We define node scoring functions based on operations applied to data related to the node and the specific problem. A SS heuristic is then implemented using a best-first search (BFS) approach, which iteratively prioritizes nodes based on their scores. The space of SS heuristics is explored through GP, with the quality of each policy assessed on a training set. The method is presented in~\cref{GP_section}.

Our approach, integrated into the SCIP solver, is benchmarked against SCIP's native method, a recent a neural network-based imitation learning algorithm~\cite{labassi_learning_2022}, and three state-of-the-art heuristics. In the first phase, we evaluate performance on three primal hard problem types: Fixed Charge Multicommodity Network Flow (FCMCNF), Maximum Satisfiability (MAXSAT), and Generalized Independent Set (GISP). Each method is evaluated on instances comparable in size to the training set as well as on larger instances. Our method achieves solving times within 2\% of the best baseline for each instance set. Compared to the SCIP, \emph{GP2S} consistently performs equally or better, with an average speedup of 11.3\%.  In the second phase, we apply \emph{GP2S} to MIPLIB 2017, training multiple GP-based SS heuristics on $50$ instances with a 10-second time limit. Seven out of ten heuristics outperform SCIP at a 150-second time limit, with some methods excelling in terms of the number of feasible solutions or optimality gap across various instance sets and time limits.  Detailed experimental results are provided in~\cref{experiments}.

The contributions of this paper are as follows:
\begin{itemize}
    \item Definition of a space of SS heuristics using a scoring function that integrates node and problem data, including existing handcrafted heuristics.
    \item Design of a GP algorithm to explore this space, aiming to improve B\&B performance on a training set of instances.
    \item When trained on a specific problem type, \emph{GP2S} performs within 2\% of the best baseline and achieves an average speedup of 11.3\% over SCIP, even on larger instances than those used for training.
    \item Evaluation on the MIPLIB 2017 instance set, contrasting with the focus on synthetically generated problems in most evaluations.
    \item On MIPLIB 2017, \emph{GP2S} surpasses SCIP in 7 out of 10 cases on 15 times more instances than those used for training, with a time limit 15 times longer, with some GP-based methods excelling in terms of feasible solutions or optimality gap on both reduced and full sets across different time limits.
\end{itemize}

\section{Background}~\label{background}

\subsubsection{SS Policies in B\&B}

A MILP can be formulated as ${\min}_{x \in N_0} \{c^T x : A x \geq b\}$, where $N_0 = \mathbb{Z}^k \times \mathbb{R}^{n-k}$, $A \in \mathbb{R}^{m \times n}$, $b \in \mathbb{R}^m$ and $c \in \mathbb{R}^n$ are vectors of constants.
The B\&B algorithm addresses this problem through the following steps:
(i) relaxation: solve the relaxed problem over $x \in N_i$ (initially $N_i = N_0$). Let $x_i=(x_i^j)_{1\leq j\leq n}$ denote the optimal solution and $z_i$ represent the objective value; 
(ii) feasibility check: if $x_i$ satisfies the integrality constraints, update the best objective value to $z^* = \min(z^*, z_i)$; if $x_i$ does not meet the integrality constraints, $z_i$ serves as a lower bound (LB); if $z_i$ exceeds $z^*$ ($z^* < z_i$), the problem is \emph{pruned};
(iii) branching: if the problem is not pruned, select a variable $x^j$ (for $1 \leq j \leq k$) that fails to meet integrality constraints; generate two subproblems: $N_{i}^{j-} = N_i$, $x^j \leq \lfloor x_i^j \rfloor$ and  $N_i^{j+} = N_i$, $x^j \geq \lfloor x_i^j \rfloor + 1$; additionally, denote $d_i$ as the total number of times a variable has been selected as a branching variable to define $N_i$, so  $d_i^{j+} = d_i+1$ and $d_i^{j-} = d_i+1$; here, $d_i$ represents the depth of the node in the tree;
(iv) exploration: continue by selecting a new subproblem from the remaining unexplored domains $N_i$ and terminate once all subproblems are either explored or pruned.
The process of selecting the next node to explore is known as the \emph{search strategy} (SS). The efficiency of finding the best-known solution $z^*$ is significantly impacted by the sequence in which nodes are explored, with faster convergence to high-quality solutions enhancing the effectiveness of subsequent pruning.
A SS policy can be characterized by two components. First, the \emph{node-comparing} method establishes a relationship (dominance or equivalence) between two nodes. Second, the \emph{node-selection} method determines which node to explore, typically selecting the highest-ranked node (defining a BFS) or through a more complex paradigm. This structure is exemplified in SCIP~\cite{bolusani_scip_2024}, which uses a similar decomposition.

\subsubsection{GP Framework}\label{genetic_programming}
An evolutionary algorithm is a population-based metaheuristic where a population of individuals is evaluated and evolved using operators to enhance their quality assessed via a fitness function. A key feature of GP is that each individual is defined as a computer program. When these individuals represent heuristics, GP explores the heuristic space, thus classifying it as a generative hyper-heuristic~\cite{dokeroglu_hyper-heuristics_2024}.
An evolutionary algorithm, including GP, follows a generic structure for defining and evolving individuals within a population. After generating the initial population, the population is refined through multiple iterations, or generations, using three primary evolutionary processes, typically in the following order: 
(i) \emph{selection}, which determines which individuals are chosen for the next processes according to their fitness and how many offspring each selected individual produces;
(ii) \emph{crossover}, which combines two or more individuals to generate new ones; and
(iii) \emph{mutation}, which modifies an individual. 
The evolutionary process concludes based on criteria such as the number of generations, elapsed time, or convergence rate, after which the best individual(s) are returned. For a more detailed understanding of evolutionary methods, refer to~\cite{talbi_metaheuristics_2009}.
In GP, individuals are often structured as trees, which simplifies the representation of mutations and crossovers; the programs are reassembled recursively~\cite{koza_hierarchical_1989}. The individuals in this structure include two main types of components: \emph{terminals}, found at the leaves of the tree and acting as the building blocks of the program, and \emph{operators}, located at non-leaf nodes, which perform operations on one or more terminals depending on their arity. Typically, all terminals are of the same type, and operators produce results of the same type.

\section{Related Work}~\label{related_works}

\subsubsection{Handcrafted SS Heuristics}

The literature on SS for B\&B includes various handcrafted heuristics employing different methods for node comparison and selection~\cite{lawler_branch-and-bound_1966, morrison_branch-and-bound_2016, tomazella_comprehensive_2020}. Typically, a node is evaluated based on features derived from the relaxation of the problem, with node comparison involving these feature-based scores. A common approach is to prioritize nodes with the smallest LB $z_i$, assuming that if the relaxed problem yields a promising solution, the original problem is likely to follow suit.
An alternative method introduced by~\cite{benichou_experiments_1971} provides an estimate of the objective function's value. For a branching variable $x^j$, its fractional part is $f_i^j = x_i^j - \lfloor x_i^j \rfloor$. The objective values of the related subdomains $N_{i}^{j-}$ and $N_{i}^{j+}$ are denoted as $z_{i}^{j-}$ and $z_{i}^{j+}$, respectively. Pseudocosts are computed as $P_{i}^{j-} = (z_{i}^{j-} - z_i)/f_i^j$ and $P_{i}^{j+} = (z_{i}^{j+} - z_i)/(1 - f_i^j)$. The metric, known as the \emph{best estimate} at node $N_i$, is given by $\text{BE}_i = z_i + \sum_{1 \leq j \leq k} \min(P_{i}^{j-} f_i^j, P_{i}^{j+} (1 - f_i^j))$. For node selection, the best node is generally chosen from a set of open nodes, with ranking determined by the node-comparing method. The set of open nodes can include all nodes for BFS, offspring nodes relative to the active node for depth-first search (DFS), or siblings for breadth-first search, with potential transitions from DFS to BFS upon reaching a leaf node.

No single method dominates the others for node evaluation or node selection. For example, BFS is theoretically more efficient but can significantly increase the open node list size, leading to scheduling and memory issues that slow down the process.~\cite{linderoth_computational_1999} evaluated 13 handcrafted SS heuristics, integrating the node-comparing and node-selection methods discussed.
These heuristics are highly interpretable and computationally efficient but are manually constructed, indicating that automating their generation could potentially enhance performance.

\subsubsection{ML for MILP}
Recent advancements have seen ML being increasingly applied to MILP, as summarized in several surveys.~\cite{lodi_learning_2017} reviews ML contributions to B\&B SS policies, focusing on approaches using imitation learning and reinforcement learning.~\cite{bengio_machine_2020} provides an in-depth survey on the intersection of ML and combinatorial optimization, exploring various integrations between the two fields and different ML learning objectives.~\cite{cappart_combinatorial_2022} discusses the applications of graph neural networks (GNN) to optimization problems. Furthermore,~\cite{scavuzzo_machine_2024} examines ML’s role in developing primal heuristics, branching strategies, SS policies, cutting planes,  and other solver configurations. 

\subsubsection{Classification Method Applied to Node Selection}

The node-comparing method can be framed as a classification problem with three potential outcomes: one node dominating another, the other node dominating, or equivalence between the nodes. 
The decision should be made based on the current state, which includes the compared nodes, the problem's features, and the current progress of the B\&B process. 
Classification models are trained on previously solved instances with the objective of identifying the dominant node that leads to faster resolution, using imitation learning on an oracle that is presumed to be the perfect solver. 
These methods are applied to obtain a fixed number of feasible solutions before switching to a handcrafted method, such as best-estimate BFS, to complete the resolution.
\cite{he_learning_2014} introduced a selection method and a pruning technique based on support vector machines and the DAgger algorithm. \cite{yilmaz_study_2021,song_learning_2018} trained a multilayer perceptron model with the same feature as~\cite{he_learning_2014}, while~\cite{labassi_learning_2022} employed a GNN. Each of these methods has its specific characteristics, but the GNN approach is generally more versatile, enabling it to handle problems of different sizes.
However, imitation learning is confined to predefined methods within the solver. Additionally, methods based on neural networks require significant computational resources for each node comparison and function as black boxes. Furthermore, these methods are only applied during the initial stages of tree exploration,  and thus cannot be considered comprehensive SS policies.



\subsubsection{GP Based on Node Scoring}

GP is also a branch of ML, though it remains relatively unexplored in combination with MILP; no works related to GP are reported in the following surveys~\cite{lodi_learning_2017, bengio_machine_2020, cappart_combinatorial_2022, scavuzzo_machine_2024}. However, there are preliminary works proposing the application of GP for SS in B\&B. In two papers~\cite{kostikas_genetic_2004, morikawa_job_2019}, GP is used to define a function that evaluates the quality of a node, enabling the construction of a SS policy by iteratively selecting the node with the highest quality.
\cite{kostikas_genetic_2004} introduced a two-phase B\&B approach. The first phase involves a traditional exploration of the search space, followed by a second phase where a GP-based method is employed to develop a new SS policy. The fitness function for this GP approach is defined based on improvements in integer infeasibilities during dives, according to the tree constructed in the first phase.~\cite{morikawa_job_2019} developed a GP scoring method using terminals specific to job shop scheduling.

\subsubsection{Positioning}

Our methodology leverages a generic hyper-heuristic GP approach akin to~\cite{kostikas_genetic_2004}, with key distinctions. Unlike their two-phase resolution process tailored to individual instances, our GP build heuristics that generalize effectively across diverse instance sets. While their implementation in GLPK omits critical details—such as the initial SS, phase-transition criteria, and GP configurations—we integrate our method with the modern SCIP solver and provide an open-source implementation. Experimentally, their approach shows limited performance gains and is at least twice as slow as contemporary state-of-the-art heuristics, highlighting the inefficiency of two-phase strategies.
Unlike the neural network-based approaches proposed by~\cite{yilmaz_study_2021,song_learning_2018,labassi_learning_2022}, our focus is on constructing an SS policy that is both computationally lightweight for node comparison and interpretable. Since our heuristic space encompasses most handcrafted heuristics, the goal is to develop a policy that is smarter than those previously discovered. This approach is classified under ``Learning to configure algorithms" with a ``multi-instance" learning objective, as defined in~\cite{bengio_machine_2020}. 


\section{GP-based Node Scoring Function}~\label{GP_section}

We now introduce \emph{GP2S}, an ML method that generate a SS policy for B\&B through GP. A policy is represented as the scoring of a B\&B node, coupled with a BFS approach that always selects the node with the best score. The scoring function has a tree structure, and we explore the space of scoring functions using the GP method. 
Firstly, we delineate the SS policy space, followed by an explanation of our choices for the evolutionary algorithm tuning.


\subsection{Definition of the SS Policy Space}
We define the space of scoring functions using standard operations on features derived from both the node and the problem. The SS is then constructed by combining these functions with a BFS-based node-selection method. Since the scoring function is real-valued, we utilize terminals of floating-point types and operators tailored for real number operations, as summarized in~\cref{operators_terminals}.
The operators are selected as standard arithmetic functions over real numbers. To integrate our approach within the SCIP solver, we employ real-type variables that are either node-specific or model-specific—commonly used attributes for characterizing nodes and models. Node-related variables vary from node to node, while model-related variables differ between instances but remain constant within a single instance. Additionally, we introduce the constant $M$, a sufficiently large positive value. This allows for the definition of multiple levels of scoring, particularly when the scoring function consists of the sum of several components, each weighted by different powers of $M$. For instance, the SS policy combining DFS with the lowest LB can be represented in this way: depth is prioritized first, and in the case of a tie, the node with the lowest LB is selected. The corresponding minimization function is expressed as $f(N_i) = z_i - M \times d_i$.
This construction significantly expands the SS policy space beyond manually defined heuristics. Given $\alpha$ as the number of 2-arity operators and $\beta$ as the number of terminals, $\beta^{2^r} \times \alpha^{2^r-1}$ perfect trees of size $r$ can be generated. As a numerical example, using the operators and terminals detailed in~\cref{operators_terminals}, approximately $10^{11}$ different trees of depth 3 can be obtained.

\begin{table}[htbp]
    \centering
    \footnotesize
    \begin{tabular}{cc}
        \begin{tabular}{cm{1.7cm}}
    \multicolumn{2}{c}{\textsc{\textbf{Operators}}}\\
    \multicolumn{1}{c}{\textbf{Sym}} & \multicolumn{1}{c}{\textbf{Meaning}} \\\hline\hline
         $+$&  Add two inputs\\ \hline
         $-$& Substract two inputs\\\hline%
         $\times$ & Multiply two inputs
         \\\hline%
         $/$&  Divide with protection: if division by $0$, return $1$%
    \end{tabular} & \begin{tabular}{ll}
    \multicolumn{2}{c}{\textsc{\textbf{Terminals}}}\\
    \multicolumn{1}{c}{\textbf{Sym}} & \multicolumn{1}{c}{\textbf{Name}} \\
    \hline\hline
    
    \multicolumn{2}{c}{\textbf{Node-related}}\\\hline    
       $d_i$  & Depth \\  
         $\text{BE}_i$ &Best estimate\\
         $z_i$ &Lower bound  \\\hline
         \multicolumn{2}{c}{\textbf{Model-related}}\\\hline
         $z_0$ & Dual bound at the root  \\
         $m$ & Number of constraints  \\
          $n$ & Number of variables  \\\hline
          \multicolumn{2}{c}{\textbf{Constant}}\\\hline
          $M$ & Big positive constant\\
    \end{tabular}
    \end{tabular}
    
    \caption{Operators and terminals.}
    \label{operators_terminals}
\end{table}

\subsection{GP Framework}\label{gp_framework}
We base our GP approach on the framework outlined in \cite{baeck_evolutionary_2000}, which includes parameters such as population size $p$, number of generations $g$, and probabilities for crossover $P_\text{mate} \in [0,1]$ and mutation $P_\text{mutate} \in [0,1]$. The process begins with the creation of an initial population of size $p$. Each generation starts with the selection phase, where individuals are chosen (with possible duplication) to maintain the population size. During the evolution phase, individuals are paired for crossover with probability $P_\text{mate}$. Following crossover, mutation occurs where each individual in the population has a probability of $P_\text{mutate}$ of undergoing mutation.

\subsubsection{Fitness Function}~\label{fitness}

A crucial aspect of defining an evolutionary algorithm is the characterization of an individual's fitness, which signifies the desired traits and is refined through the evolutionary process, serving as a discriminating factor during selection. In our case, we may aim to define an SS policy for either complete or incomplete instance resolution (typically under a time limit), which leads to two distinct methods for evaluating the policy's performance on an instance. For each performance measure defined on the training set instances, the fitness function is based on the 1-shifted geometric mean, a standard metric for assessing average performance on a benchmark~\cite{achterberg_constraint_2007}. 
For the complete resolution objective, as examined in the first part of the simulation, the focus is on solving instances as quickly as possible, so solving time is used as the criterion. In the case of incomplete resolution, explored in the second part of the simulation, the performance of the best-found solution (if any) is assessed using the \emph{optimality gap}. This gap measures the relative difference between the best integer solution and the lowest LB among unexplored nodes. The lowest LB is defined as $\text{Best LB} = \min_{i, N_i \text{ unexplored}}(z_i)$, and the gap is calculated as $\text{Gap} = \left| (z^* - \text{Best LB})/(\min(z^*, \text{Best LB})) \right|$. If the optimal solution is found, then $\text{Gap}=0$. By convention, if no solution is found, the optimality gap is set to $1\mathrm{e}+20$ (as returned by the SCIP solver). This fitness function is structured to first minimize the number of infeasible solutions and then to speed up convergence toward a better solution.





\subsubsection{GP Population Initialization}
Initially, a population of $ p $ individuals, represented as trees, is generated. The generation of an individual begins at the root node, whose nature is determined by selecting uniformly at random from the set of operators and terminals. If an operator is chosen, its child nodes are generated, and the process continues. If a terminal is selected, the branch is completed. Each branch is constrained to have a minimum depth of $ D_\text{init-min} $ and a maximum depth of $ D_\text{init-max} $.

\subsubsection{Selection}

\begin{figure}[htbp]
\centering
\includegraphics[scale=0.24]{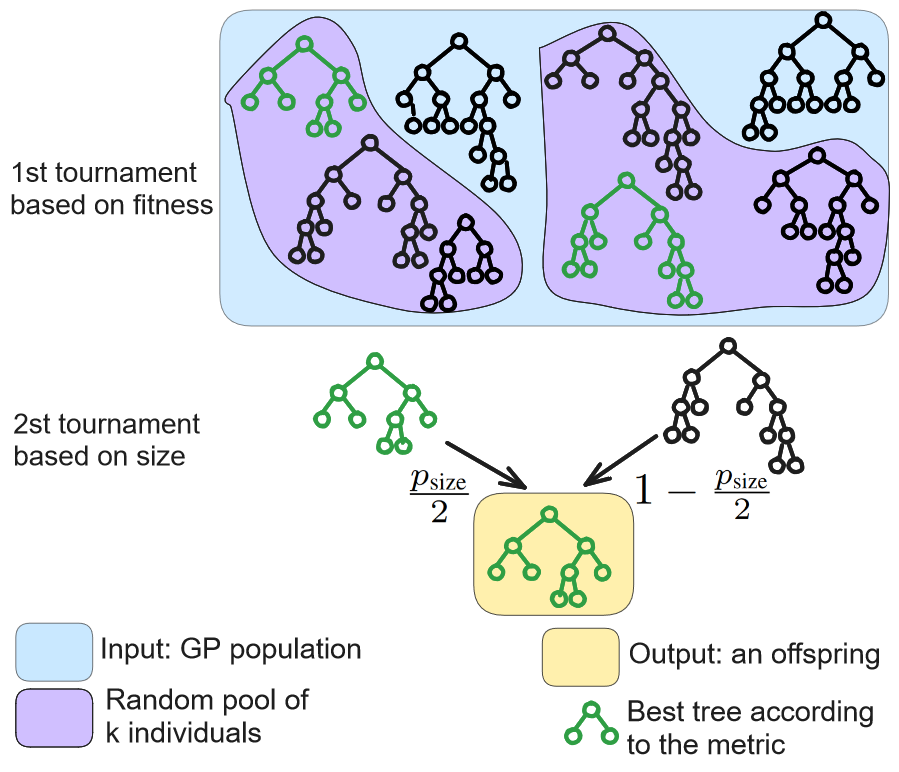}
\caption{Selection process based on double tournament: this step is repeated $ p $ times to generate $ p $ offspring.}
\label{selection}
\end{figure}

For selection, we use a double tournament method as defined in~\cite{luke_fighting_2002}, designed to avoid convergence towards excessively deep trees. Deep trees can incur higher costs during node score calculations and are particularly prone to overfitting, whereas simpler formulations tend to be more general. In the first tournament, of size $ k $, we randomly select $ k $ individuals from the GP population and retain the best one based on one of the previously defined fitness criteria.  After performing two fitness-based tournaments, the two winning candidates proceed to a second round, where the deciding factor is the total number of nodes in the tree. for this second round, to favor the selection of smaller trees, the smaller tree has a probability $\frac{P_\text{size}}{2}$ of being selected, where $ P_\text{size} \in [1,2] $. An illustration of the double tournament selection process is presented in~\cref{selection}.

\subsubsection{Crossover and Mutation}

\begin{figure}[htbp]
\centering
\includegraphics[scale=0.24]{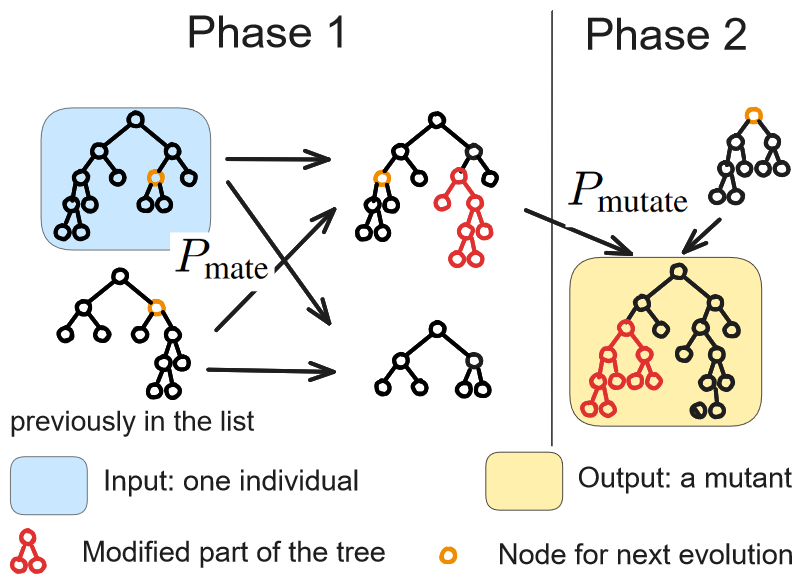}
\caption{Crossover and mutation processes: the whole population completes phase 1, then moves to phase 2.}
\label{evolution process}
\end{figure}

After the selection process comes the crossover, then mutation.
If a crossover operation is performed for an individual (with probability $ P_\text{mate} $), another individual (typically the previous one in the selection list) is chosen from the population. A one-point crossover is then executed by selecting a node in each tree and swapping the subtrees rooted at these nodes, thereby modifying both offspring. If no crossover occurs, the individual remains unchanged. Next, we perform the mutation process on the modified population. With probability $ P_\text{mutate} $ (otherwise unchanged with probability $ 1 - P_\text{mutate} $), a uniform mutation is applied: the new tree is defined with leaves bounded by the limits $ D_\text{mut-min} $ and $ D_\text{mut-max} $. A node is chosen randomly such that the new tree grafts at this point, replacing the previous subtree. 
The crossover and mutation processes are summarized in~\cref{evolution process}.

\section{Experiments}~\label{experiments}

\begin{table*}[htbp]
\centering
\footnotesize
\begin{tabular}{l|cc|cc|cc}
\textsc{\textbf{PROBLEM}} & \multicolumn{2}{c}{\textsc{\textbf{FCMCNF}}} & \multicolumn{2}{c}{\textsc{\textbf{MAXSAT}}} & \multicolumn{2}{c}{\textsc{\textbf{GISP}}} \\\hline
\textsc{Partition} & \textsc{Test} & \textsc{Transfer}& \textsc{Test} & \textsc{Transfer}& \textsc{Test} & \textsc{Transfer}  \\\hline
BE DFS& \cellcolor[cmyk]{1.0,0.0,1,0.0}$2.9$$\pm$$1.5$& \cellcolor[cmyk]{1.0,0.0,1,0.0}$17.4$$\pm$$1.9$& \cellcolor[cmyk]{0.61,0.39,1,0.1}$5.2$$\pm$$1.9$& \cellcolor[cmyk]{0.8,0.2,1,0.05}$9.6$$\pm$$1.9$& \cellcolor[cmyk]{0.0,1.0,1,0.25}$3.4$$\pm$$1.5$& \cellcolor[cmyk]{0.0,1.0,1,0.25}$34.3$$\pm$$1.7$\\
BE BFS& \cellcolor[cmyk]{0.91,0.09,1,0.02}$3.0$$\pm$$1.5$& \cellcolor[cmyk]{0.83,0.17,1,0.04}$18.6$$\pm$$1.9$& \cellcolor[cmyk]{0.0,1.0,1,0.25}$6.3$$\pm$$1.8$& \cellcolor[cmyk]{0.0,1.0,1,0.25}$12.6$$\pm$$1.8$& \cellcolor[cmyk]{0.69,0.31,1,0.08}$2.7$$\pm$$1.6$& \cellcolor[cmyk]{0.37,0.63,1,0.16}$26.2$$\pm$$1.7$\\
LB BFS& \cellcolor[cmyk]{0.91,0.09,1,0.02}$3.0$$\pm$$1.5$& \cellcolor[cmyk]{0.63,0.37,1,0.09}$20.0$$\pm$$2.0$& \cellcolor[cmyk]{0.94,0.06,1,0.01}$4.6$$\pm$$1.8$& \cellcolor[cmyk]{0.63,0.37,1,0.09}$10.2$$\pm$$2.1$& \cellcolor[cmyk]{0.48,0.52,1,0.13}$2.9$$\pm$$1.6$& \cellcolor[cmyk]{0.44,0.56,1,0.14}$25.6$$\pm$$1.5$\\
GNN 2 dives& \cellcolor[cmyk]{0.22,0.78,1,0.19}$3.8$$\pm$$1.4$& \cellcolor[cmyk]{0.63,0.37,1,0.09}$20.0$$\pm$$1.9$& \cellcolor[cmyk]{0.33,0.67,1,0.17}$5.7$$\pm$$1.8$& \cellcolor[cmyk]{0.41,0.59,1,0.15}$11.0$$\pm$$2.1$& \cellcolor[cmyk]{0.0,1.0,1,0.25}$3.6$$\pm$$1.5$& \cellcolor[cmyk]{0.44,0.56,1,0.14}$25.6$$\pm$$1.5$\\
GNN full& \cellcolor[cmyk]{0.05,0.95,1,0.24}$4.0$$\pm$$1.6$& \cellcolor[cmyk]{0.0,1.0,1,0.25}$27.3$$\pm$$2.4$& \cellcolor[cmyk]{0.78,0.22,1,0.06}$4.9$$\pm$$1.9$& \cellcolor[cmyk]{0.55,0.45,1,0.11}$10.5$$\pm$$2.3$& \cellcolor[cmyk]{0.0,1.0,1,0.25}$3.5$$\pm$$1.6$& \cellcolor[cmyk]{0.0,1.0,1,0.25}$31.3$$\pm$$1.5$\\
SCIP& \cellcolor[cmyk]{0.66,0.34,1,0.09}$3.3$$\pm$$1.5$& \cellcolor[cmyk]{0.63,0.37,1,0.09}$20.0$$\pm$$1.9$& \cellcolor[cmyk]{0.28,0.72,1,0.18}$5.8$$\pm$$1.7$& \cellcolor[cmyk]{0.61,0.39,1,0.1}$10.3$$\pm$$1.8$& \cellcolor[cmyk]{1.0,0.0,1,0.0}$2.4$$\pm$$1.6$& \cellcolor[cmyk]{1.0,0.0,1,0.0}$20.9$$\pm$$1.6$\\
GP2S& \cellcolor[cmyk]{0.91,0.09,1,0.02}$3.0$$\pm$$1.5$& \cellcolor[cmyk]{0.96,0.04,1,0.01}$17.7$$\pm$$1.9$& \cellcolor[cmyk]{1.0,0.0,1,0.0}$4.5$$\pm$$1.8$& \cellcolor[cmyk]{1.0,0.0,1,0.0}$8.9$$\pm$$2.1$& \cellcolor[cmyk]{1.0,0.0,1,0.0}$2.4$$\pm$$1.6$& \cellcolor[cmyk]{1.0,0.0,1,0.0}$20.9$$\pm$$1.6$\\

\end{tabular}
\caption{Solving time with standard deviation (1-shifted geometric mean and geometric standard deviation) for our proposal and baselines on three problem types, for both training-similar (test) and larger instances (transfer).
 }
\label{simulation_artificial}
\end{table*}

Our experiments are conducted in two phases. In the first phase, \emph{GP2S} is evaluated on three synthetic problem types, testing performance and generalizability on instances similar in size to the training set and larger instances. In the second phase, \emph{GP2S} heuristics are generated from a subset of MIPLIB 2017 instances under a solving time limit and evaluated on the full dataset with extended time limits. This phase examines the heuristic's generalization to a heterogeneous dataset and efficiency within constrained time. All simulations use the SCIP solver~\cite{bolusani_scip_2024,maher_pyscipopt_2016}, with complete code available on the Git repository: \url{https://gitlab.com/uniluxembourg/snt/pcog/ultrabo/search-strategy-generation-for-branch-and-bound-using-genetic-programming.git}. Detailed integration of our method into SCIP is also provided within the framework.

As we are evaluating performance within SCIP, we use the default SCIP SS policy as the primary baseline, making no modifications to the solving core (denoted as \emph{SCIP} in the performance evaluation table).  We further compare our approach to several manually crafted heuristics. Specifically, we include the method that ranks nodes by their LB in a BFS (\emph{LB BFS}), as well as the best estimate in BFS (\emph{BE BFS}) and DFS (\emph{BE DFS}). Additionally, for the first part of the experiment, we include a recent GNN-based SS method~\cite{labassi_learning_2022}. The original GNN SS policy is implemented in two phases: the trained GNN is used until two solutions are found, after which the policy switches to \emph{BE BFS}. For comparison, we include this approach (\emph{GNN 2 dives}), as well as the variant where the trained GNN-based node-comparison method is applied throughout the entire resolution process (\emph{GNN full}). This method requires the complete resolution of training instances so that the GNN learns to replicate decisions leading to the best solutions. However, because complete B\&B resolution is not guaranteed for MIPLIB 2017 instances, this method is not included in the comparisons for the second part of the experiment. 

\emph{GP2S} is a parametric approach that requires defining evolutionary features. We have chosen classical parameters that are widely used in the development of such methods~\cite{ahvanooey_survey_2019,luke_fighting_2002,duflo_gp_2019}: the probabilities for crossover and mutation are set to $P_\text{mate}=0.9$ and $P_\text{mutate}=0.1$; the depth of population initialization is restricted by minimum and maximum limits of $D_\text{init-min}=1$ and $D_\text{init-max}=17$; the depth of branches generated during mutation is bounded by $D_\text{mut-min}=1$ and $D_\text{mut-max}=5$; the parameter related to the selection of the smallest tree size is set to $P_\text{size} = 1.2$. The parameters governing the number of individuals and generations are detailed in each part of the experiment, as these factors greatly influence the computational cost of the solution. To visually compare the performance of each baseline across different sets of instances and evaluation metrics, we color the cells in the performance table: {green} indicates the best results, {red} marks results that are more than $40\%$ worse than the optimum, and a color gradient is applied for intermediate results.


\subsection{Solving Time for Specific Types of Problem}~\label{specific_problems}

\subsubsection{Types of Problem Solved}\label{bench}

We benchmark our method against the SS policy from~\cite{labassi_learning_2022} using similar experimental settings, including problem types and instance sizes, that enable indirect comparisons with methods evaluated in their study~\cite{he_learning_2014,yilmaz_study_2021,song_learning_2018}. We evaluate three challenging primal problem types, each with 50 instances: a training set for development, a test set with similar instances for effectiveness, and a transfer set with larger instances for generalizability. The problem types are: 
(i) FCMCNF~\cite{hewitt_combining_2009}, with $n=15$ nodes for training and testing, and $n=20$ for transfer, with $m=1.5n$ commodities~\cite{chmiela_learning_2021}; 
(ii) MAXSAT, with $n \in [60,70]$ for training and testing, and $n \in [70,80]$ for transfer~\cite{bejar_generating_2009}; 
(iii) GISP, with $n \in [60,70]$ for training and testing, and $n \in [80,100]$ for transfer~\cite{chmiela_learning_2021}. 
All problems are defined on Erdos-Rényi random graphs~\cite{erdos_evolution_1961} with $p = 0.3$ for FCMCNF and $p = 0.6$ for MAXSAT and GISP. Each instance is solved to optimality.

\subsubsection{Hardware and GP2S Setup}

In the first phase, a neural network-based method is used as the baseline, requiring GPU-equipped hardware. Each run, involving solving an instance with a baseline SS policy, is executed on a node with 7 CPU cores, 192 GB of RAM, and an NVIDIA V100 GPU with 16 GB of RAM.

For the \emph{GP2S} setup, the method uses 50 individuals over 50 generations with a tournament size of $k = 5$. The fitness function, as detailed in~\cref{gp_framework}, evaluates the GP-based SS policy using the geometric mean solving time across 50 training instances for each problem. GP convergence plots and scoring functions are presented in~\cref{GP_three_problems}.

\begin{table*}[htbp]
\centering
\footnotesize
\begin{tabular}{@{}l|cc|cc|cc||cc|cc|cc}
& \multicolumn{6}{c}{REDUCED MIPLIB 2017: 166 instances}&\multicolumn{6}{c}{ALL MIPLIB 2017: 758 instances}\\
&\multicolumn{2}{c}{10 seconds}& \multicolumn{2}{c}{50 seconds} &\multicolumn{2}{c}{150 seconds}&\multicolumn{2}{c}{10 seconds}& \multicolumn{2}{c}{50 seconds} &\multicolumn{2}{c}{150 seconds}\\ \hline
&\textsc{Inf} & \textsc{Gap}& \textsc{Inf} & \textsc{Gap}& \textsc{Inf} & \textsc{Gap}& \textsc{Inf} & \textsc{Gap}& \textsc{Inf} & \textsc{Gap}& \textsc{Inf} & \textsc{Gap}\\
BE BFS & \cellcolor[cmyk]{0.92,0.08,1,0.02}$ 34$ & \cellcolor[cmyk]{0.96,0.04,1,0.01}$0.72$$\pm$$4.0$ & \cellcolor[cmyk]{0.0,1.0,1,0.25}$ 33$ & \cellcolor[cmyk]{0.0,1.0,1,0.25}$0.62$$\pm$$3.5$ & \cellcolor[cmyk]{0.0,1.0,1,0.25}$ 26$ & \cellcolor[cmyk]{0.0,1.0,1,0.25}$0.45$$\pm$$3.8$ & \cellcolor[cmyk]{0.58,0.42,1,0.1}$ 455$ & \cellcolor[cmyk]{0.0,1.0,1,0.25}$2.08$$\pm$$5.8$ & \cellcolor[cmyk]{0.95,0.05,1,0.01}$ 289$ & \cellcolor[cmyk]{0.97,0.03,1,0.01}$0.75$$\pm$$4.4$ & \cellcolor[cmyk]{0.82,0.18,1,0.05}$ 249$ & \cellcolor[cmyk]{0.89,0.11,1,0.03}$0.69$$\pm$$4.4$\\
BE DFS & \cellcolor[cmyk]{0.92,0.08,1,0.02}$ 34$ & \cellcolor[cmyk]{0.86,0.14,1,0.04}$0.75$$\pm$$4.1$ & \cellcolor[cmyk]{0.32,0.68,1,0.17}$ 28$ & \cellcolor[cmyk]{0.68,0.32,1,0.08}$0.44$$\pm$$3.9$ & \cellcolor[cmyk]{0.22,0.78,1,0.2}$ 21$ & \cellcolor[cmyk]{0.55,0.45,1,0.11}$0.33$$\pm$$4.5$ & \cellcolor[cmyk]{0.99,0.01,1,0.0}$ 391$ & \cellcolor[cmyk]{0.85,0.15,1,0.04}$1.06$$\pm$$4.5$ & \cellcolor[cmyk]{0.88,0.12,1,0.03}$ 297$ & \cellcolor[cmyk]{0.9,0.1,1,0.03}$0.77$$\pm$$4.4$ & \cellcolor[cmyk]{0.91,0.09,1,0.02}$ 240$ & \cellcolor[cmyk]{0.77,0.23,1,0.06}$0.72$$\pm$$4.5$\\
LB BFS & \cellcolor[cmyk]{0.0,1.0,1,0.25}$ 51$ & \cellcolor[cmyk]{0.0,1.0,1,0.25}$1.26$$\pm$$4.4$ & \cellcolor[cmyk]{0.09,0.91,1,0.23}$ 30$ & \cellcolor[cmyk]{1.0,0.0,1,0.0}$0.39$$\pm$$4.1$ & \cellcolor[cmyk]{0.0,1.0,1,0.25}$ 30$ & \cellcolor[cmyk]{0.0,1.0,1,0.25}$0.4$$\pm$$4.0$ & \cellcolor[cmyk]{0.95,0.05,1,0.01}$ 398$ & \cellcolor[cmyk]{0.92,0.08,1,0.02}$1.03$$\pm$$4.3$ & \cellcolor[cmyk]{0.87,0.13,1,0.03}$ 298$ & \cellcolor[cmyk]{1.0,0.0,1,0.0}$0.74$$\pm$$4.4$ & \cellcolor[cmyk]{0.22,0.78,1,0.19}$ 304$ & \cellcolor[cmyk]{0.0,1.0,1,0.25}$1.14$$\pm$$4.8$\\
SCIP & \cellcolor[cmyk]{0.62,0.38,1,0.09}$ 38$ & \cellcolor[cmyk]{0.82,0.18,1,0.04}$0.76$$\pm$$4.1$ & \cellcolor[cmyk]{0.55,0.45,1,0.11}$ 26$ & \cellcolor[cmyk]{0.87,0.13,1,0.03}$0.41$$\pm$$4.0$ & \cellcolor[cmyk]{0.0,1.0,1,0.25}$ 26$ & \cellcolor[cmyk]{0.0,1.0,1,0.25}$0.41$$\pm$$4.1$ & \cellcolor[cmyk]{0.98,0.02,1,0.0}$ 393$ & \cellcolor[cmyk]{0.85,0.15,1,0.04}$1.06$$\pm$$4.5$ & \cellcolor[cmyk]{0.95,0.05,1,0.01}$ 289$ & \cellcolor[cmyk]{0.93,0.07,1,0.02}$0.76$$\pm$$4.4$ & \cellcolor[cmyk]{0.34,0.66,1,0.16}$ 293$ & \cellcolor[cmyk]{0.0,1.0,1,0.25}$1.14$$\pm$$4.8$\\
GP2S 2 & \cellcolor[cmyk]{0.85,0.15,1,0.04}$ 35$ & \cellcolor[cmyk]{0.93,0.07,1,0.02}$0.73$$\pm$$4.0$ & \cellcolor[cmyk]{0.43,0.57,1,0.14}$ 27$ & \cellcolor[cmyk]{0.81,0.19,1,0.05}$0.42$$\pm$$3.9$ & \cellcolor[cmyk]{0.0,1.0,1,0.25}$ 26$ & \cellcolor[cmyk]{0.46,0.54,1,0.13}$0.34$$\pm$$4.3$ & \cellcolor[cmyk]{0.59,0.41,1,0.1}$ 454$ & \cellcolor[cmyk]{0.0,1.0,1,0.25}$2.08$$\pm$$5.8$ & \cellcolor[cmyk]{0.88,0.12,1,0.03}$ 297$ & \cellcolor[cmyk]{0.97,0.03,1,0.01}$0.75$$\pm$$4.4$ & \cellcolor[cmyk]{0.76,0.24,1,0.06}$ 254$ & \cellcolor[cmyk]{1.0,0.0,1,0.0}$0.66$$\pm$$3.9$\\
GP2S 4 & \cellcolor[cmyk]{0.0,1.0,1,0.25}$ 49$ & \cellcolor[cmyk]{0.0,1.0,1,0.25}$1.25$$\pm$$4.4$ & \cellcolor[cmyk]{0.09,0.91,1,0.23}$ 30$ & \cellcolor[cmyk]{0.49,0.51,1,0.13}$0.47$$\pm$$3.8$ & \cellcolor[cmyk]{0.0,1.0,1,0.25}$ 30$ & \cellcolor[cmyk]{0.11,0.89,1,0.22}$0.38$$\pm$$4.2$ & \cellcolor[cmyk]{0.97,0.03,1,0.01}$ 394$ & \cellcolor[cmyk]{1.0,0.0,1,0.0}$1.0$$\pm$$4.3$ & \cellcolor[cmyk]{0.86,0.14,1,0.04}$ 299$ & \cellcolor[cmyk]{0.93,0.07,1,0.02}$0.76$$\pm$$4.3$ & \cellcolor[cmyk]{0.22,0.78,1,0.19}$ 304$ & \cellcolor[cmyk]{0.0,1.0,1,0.25}$1.16$$\pm$$4.7$\\
GP2S 6 & \cellcolor[cmyk]{0.92,0.08,1,0.02}$ 34$ & \cellcolor[cmyk]{1.0,0.0,1,0.0}$0.71$$\pm$$4.1$ & \cellcolor[cmyk]{0.43,0.57,1,0.14}$ 27$ & \cellcolor[cmyk]{0.68,0.32,1,0.08}$0.44$$\pm$$3.9$ & \cellcolor[cmyk]{0.0,1.0,1,0.25}$ 25$ & \cellcolor[cmyk]{0.29,0.71,1,0.18}$0.36$$\pm$$4.2$ & \cellcolor[cmyk]{0.99,0.01,1,0.0}$ 392$ & \cellcolor[cmyk]{0.9,0.1,1,0.03}$1.04$$\pm$$4.5$ & \cellcolor[cmyk]{0.88,0.12,1,0.03}$ 297$ & \cellcolor[cmyk]{0.9,0.1,1,0.03}$0.77$$\pm$$4.3$ & \cellcolor[cmyk]{0.77,0.23,1,0.06}$ 253$ & \cellcolor[cmyk]{0.96,0.04,1,0.01}$0.67$$\pm$$3.9$\\
GP2S 8 & \cellcolor[cmyk]{1.0,0.0,1,0.0}$ 33$ & \cellcolor[cmyk]{0.82,0.18,1,0.04}$0.76$$\pm$$4.1$ & \cellcolor[cmyk]{0.77,0.23,1,0.06}$ 24$ & \cellcolor[cmyk]{0.49,0.51,1,0.13}$0.47$$\pm$$3.8$ & \cellcolor[cmyk]{0.22,0.78,1,0.2}$ 21$ & \cellcolor[cmyk]{0.0,1.0,1,0.25}$0.42$$\pm$$3.9$ & \cellcolor[cmyk]{1.0,0.0,1,0.0}$ 390$ & \cellcolor[cmyk]{0.82,0.18,1,0.04}$1.07$$\pm$$4.5$ & \cellcolor[cmyk]{0.98,0.02,1,0.0}$ 285$ & \cellcolor[cmyk]{0.83,0.17,1,0.04}$0.79$$\pm$$4.3$ & \cellcolor[cmyk]{0.37,0.63,1,0.16}$ 290$ & \cellcolor[cmyk]{0.0,1.0,1,0.25}$1.19$$\pm$$4.7$\\
GP2S 10 & \cellcolor[cmyk]{0.7,0.3,1,0.08}$ 37$ & \cellcolor[cmyk]{0.93,0.07,1,0.02}$0.73$$\pm$$4.1$ & \cellcolor[cmyk]{0.32,0.68,1,0.17}$ 28$ & \cellcolor[cmyk]{0.68,0.32,1,0.08}$0.44$$\pm$$3.9$ & \cellcolor[cmyk]{0.0,1.0,1,0.25}$ 24$ & \cellcolor[cmyk]{0.38,0.62,1,0.16}$0.35$$\pm$$4.3$ & \cellcolor[cmyk]{0.97,0.03,1,0.01}$ 395$ & \cellcolor[cmyk]{0.92,0.08,1,0.02}$1.03$$\pm$$4.4$ & \cellcolor[cmyk]{0.87,0.13,1,0.03}$ 298$ & \cellcolor[cmyk]{0.9,0.1,1,0.03}$0.77$$\pm$$4.4$ & \cellcolor[cmyk]{0.82,0.18,1,0.05}$ 249$ & \cellcolor[cmyk]{0.96,0.04,1,0.01}$0.67$$\pm$$4.0$\\
GP2S 12 & \cellcolor[cmyk]{0.85,0.15,1,0.04}$ 35$ & \cellcolor[cmyk]{0.86,0.14,1,0.04}$0.75$$\pm$$4.0$ & \cellcolor[cmyk]{1.0,0.0,1,0.0}$ 22$ & \cellcolor[cmyk]{0.36,0.64,1,0.16}$0.49$$\pm$$3.8$ & \cellcolor[cmyk]{1.0,0.0,1,0.0}$ 16$ & \cellcolor[cmyk]{0.0,1.0,1,0.25}$0.41$$\pm$$4.0$ & \cellcolor[cmyk]{0.98,0.02,1,0.0}$ 393$ & \cellcolor[cmyk]{0.9,0.1,1,0.03}$1.04$$\pm$$4.3$ & \cellcolor[cmyk]{1.0,0.0,1,0.0}$ 283$ & \cellcolor[cmyk]{0.86,0.14,1,0.03}$0.78$$\pm$$4.3$ & \cellcolor[cmyk]{1.0,0.0,1,0.0}$ 232$ & \cellcolor[cmyk]{0.66,0.34,1,0.09}$0.75$$\pm$$4.3$\\
GP2S 14 & \cellcolor[cmyk]{0.77,0.23,1,0.06}$ 36$ & \cellcolor[cmyk]{0.82,0.18,1,0.04}$0.76$$\pm$$4.1$ & \cellcolor[cmyk]{0.2,0.8,1,0.2}$ 29$ & \cellcolor[cmyk]{0.42,0.58,1,0.14}$0.48$$\pm$$3.8$ & \cellcolor[cmyk]{0.0,1.0,1,0.25}$ 29$ & \cellcolor[cmyk]{0.0,1.0,1,0.25}$0.5$$\pm$$3.7$ & \cellcolor[cmyk]{0.99,0.01,1,0.0}$ 392$ & \cellcolor[cmyk]{0.87,0.13,1,0.03}$1.05$$\pm$$4.4$ & \cellcolor[cmyk]{0.88,0.12,1,0.03}$ 297$ & \cellcolor[cmyk]{0.8,0.2,1,0.05}$0.8$$\pm$$4.3$ & \cellcolor[cmyk]{0.8,0.2,1,0.05}$ 251$ & \cellcolor[cmyk]{0.66,0.34,1,0.09}$0.75$$\pm$$4.4$\\
GP2S 16 & \cellcolor[cmyk]{0.85,0.15,1,0.04}$ 35$ & \cellcolor[cmyk]{0.96,0.04,1,0.01}$0.72$$\pm$$4.1$ & \cellcolor[cmyk]{0.09,0.91,1,0.23}$ 30$ & \cellcolor[cmyk]{1.0,0.0,1,0.0}$0.39$$\pm$$4.1$ & \cellcolor[cmyk]{0.0,1.0,1,0.25}$ 29$ & \cellcolor[cmyk]{1.0,0.0,1,0.0}$0.28$$\pm$$5.0$ & \cellcolor[cmyk]{0.62,0.38,1,0.1}$ 450$ & \cellcolor[cmyk]{0.0,1.0,1,0.25}$2.07$$\pm$$5.8$ & \cellcolor[cmyk]{0.85,0.15,1,0.04}$ 300$ & \cellcolor[cmyk]{0.97,0.03,1,0.01}$0.75$$\pm$$4.4$ & \cellcolor[cmyk]{0.85,0.15,1,0.04}$ 246$ & \cellcolor[cmyk]{0.92,0.08,1,0.02}$0.68$$\pm$$4.4$\\
GP2S 18 & \cellcolor[cmyk]{0.77,0.23,1,0.06}$ 36$ & \cellcolor[cmyk]{0.79,0.21,1,0.05}$0.77$$\pm$$4.1$ & \cellcolor[cmyk]{0.66,0.34,1,0.09}$ 25$ & \cellcolor[cmyk]{0.42,0.58,1,0.14}$0.48$$\pm$$3.8$ & \cellcolor[cmyk]{0.0,1.0,1,0.25}$ 24$ & \cellcolor[cmyk]{0.02,0.98,1,0.25}$0.39$$\pm$$4.2$ & \cellcolor[cmyk]{0.97,0.03,1,0.01}$ 394$ & \cellcolor[cmyk]{0.82,0.18,1,0.04}$1.07$$\pm$$4.5$ & \cellcolor[cmyk]{0.3,0.7,1,0.17}$ 362$ & \cellcolor[cmyk]{0.0,1.0,1,0.25}$1.48$$\pm$$5.1$ & \cellcolor[cmyk]{0.31,0.69,1,0.17}$ 296$ & \cellcolor[cmyk]{0.0,1.0,1,0.25}$1.18$$\pm$$4.7$\\
GP2S 20 & \cellcolor[cmyk]{0.7,0.3,1,0.08}$ 37$ & \cellcolor[cmyk]{0.86,0.14,1,0.04}$0.75$$\pm$$4.0$ & \cellcolor[cmyk]{0.43,0.57,1,0.14}$ 27$ & \cellcolor[cmyk]{0.29,0.71,1,0.18}$0.5$$\pm$$3.8$ & \cellcolor[cmyk]{0.22,0.78,1,0.2}$ 21$ & \cellcolor[cmyk]{0.11,0.89,1,0.22}$0.38$$\pm$$4.2$ & \cellcolor[cmyk]{0.62,0.38,1,0.1}$ 450$ & \cellcolor[cmyk]{0.0,1.0,1,0.25}$2.09$$\pm$$5.8$ & \cellcolor[cmyk]{0.99,0.01,1,0.0}$ 284$ & \cellcolor[cmyk]{0.9,0.1,1,0.03}$0.77$$\pm$$4.4$ & \cellcolor[cmyk]{0.96,0.04,1,0.01}$ 236$ & \cellcolor[cmyk]{0.85,0.15,1,0.04}$0.7$$\pm$$4.4$\\
\end{tabular}
\caption{
Number of instances with no feasible solution (Inf) and geometric mean gap (Gap) for instances where all SS baselines found a solution. GP-based functions are labeled as GP2S+seed. Results are shown for reduced and full MIPLIB 2017 sets at time limits of $10$, $50$, and $150$ seconds.
}
\label{simulation_MIPLIB}
\end{table*}

\subsubsection{Evaluation and Discussion}
For each method, problem type, and instance size, we record the solving times across all instances. The average solving times, calculated using the 1-shifted geometric mean, are presented in~\cref{simulation_artificial}.

Among the handcrafted heuristics, none demonstrate consistent performance across all studied problem types. Each heuristic tends to be relatively effective on two problems but slower on the third. For the GNN-based methods, \emph{GNN 2 dives} is preferable, as the \emph{GNN full} approach can significantly slow down solving times, particularly for the FCMCNF transfer instances. The \emph{GNN 2 dives} method offers stable results for the transfer partition, avoiding extreme performance deviations, making it a more reliable SS policy compared to handcrafted heuristics. Similar results are observed for SCIP, which achieves notably better performance on GISP problems.
Regarding our proposed method, \emph{GP2S}, the solving time is at most $2\%$ slower than the best SS baseline. It either surpasses or matches SCIP’s performance across all categories, achieving an average speedup of $11.3\%$ across the six categories. Thus, \emph{GP2S} outperforms all other SS policies considered when trained on a specific problem type. It is important to note that solving an instance takes several seconds, making the total time for the GP algorithm to develop a SS policy for one problem $O(50^3)$ seconds, which translates to several days of computation. This computational effort is a one-time investment and is beneficial, as it significantly accelerates resolution for problems of such types compared to the other baselines. Furthermore, as our simulation results show, \emph{GP2S} can be trained on simpler problems while maintaining high efficiency on more complex ones.

\subsection{Time-Limited Performance on MIPLIB2017}~\label{miplib_sec}

\subsubsection{Partitions and Time Limits}
We conducted experiments using the MIPLIB 2017 collection~\cite{gleixner_miplib_2021}, excluding instances unlikely to yield meaningful results under time constraints. We removed instances tagged as \emph{feasibility, numerics, infeasible, no solution}, as well as those with presolve times exceeding 300 seconds, those unsolved within 600 seconds, or solved at the root, resulting in 758 instances. GP methods were trained with a 10-second solving time limit. To focus on informative cases, we selected a subset of 166 instances where SCIP, in standalone mode, explores more than one node within 10 seconds. Experiments were conducted with time limits ranging from 10 seconds to 150 seconds for both subsets.

\subsubsection{Hardware and GP2S Setup}
The method in~\cite{labassi_learning_2022} is not considered in this phase, as it requires a training set of pre-solved instances, which is not available for all MIPLIB 2017 instances. A CPU-only architecture is used, with a dedicated node featuring 28 cores and 128 GB of RAM to solve instances using SS baseline policies.

From the reduced set, we randomly select 50 instances with a 10-second time limit for the GP process. To account for the heterogeneity of MIPLIB 2017, we perform multiple independent GP runs to vary the training set through different seeds. To manage computational demands, we set the population size to 20, limit generations to 20, and use a fitness tournament size of $k=3$. The fitness function, detailed in~\cref{gp_framework}, uses the 1-shifted geometric mean gap (assigning $1\mathrm{e}+20$ when no integer solution is found). The resulting scoring functions are presented in~\cref{GP_MIPLIB}.


\subsubsection{Evaluation and Discussion}
To assess solution quality, we use two metrics: the number of instances with no feasible solution (where $z^* = +\infty$) and the 1-shifted geometric mean optimality gap for instances where all SS baselines found a feasible solution. Results for 10 GP methods are presented with time limits of $\{10, 50, 100\}$ in~\cref{simulation_MIPLIB}. 

No single method consistently outperforms all others across both metrics for any given time limit. However, \emph{GP2S} surpasses SCIP in at least 7 out of 10 instances, both in terms of infeasibility and optimality gap, for both the reduced and full sets at a 150-second time limit. The handcrafted heuristics generally perform well, with \emph{BE DFS} showing particularly stable results, never performing worse than 32\% relative to the best performance, making it the top method in this regard. Some GP-based methods demonstrate remarkable effectiveness: \emph{GP2S 12} has a number of unresolved instances only 5\% higher than the best-performing solution, while \emph{GP2S 16}, except for the entire MIPLIB at a 10-second time limit, achieves an optimality gap within 1\% of the best-known solution; these methods are the best according to these metrics. Given that GP-based SS policies are trained on a small pool of diverse instances, their performance can be highly variable. Yet, despite being automatically generated in just a few days of computation without any assumptions, and tested on 15 times more instances with a time limit up to 15 times longer, they perform very well compared to expert-engineered methods refined over years.

\section{Conclusion and Perspectives}~\label{conclusion}
In this paper, we have examined the SS component in B\&B, a crucial factor affecting the solving time of MILP solvers. We presented \emph{GP2S}, a method that automatically generates a heuristic from a training set, implemented in the open-source solver SCIP. We trained \emph{GP2S} on well-known synthetic problem types and demonstrated its significant superiority over the baselines, even when tested on instances larger than those used in training. When applied to the MIPLIB 2017 collection, \emph{GP2S} generated a variety of heuristics based on different training instance pools. Generally, these heuristics outperformed SCIP, with some methods surpassing all baseline approaches. Given the complexity of generating a single heuristic for all MILP problems—due to challenges such as defining the training pool and achieving generalizability—a promising direction for future work would be to define multiple problem subsets and generate a heuristic tailored to each specific subset. Another significant highlighted of this paper is the effective use of the GP paradigm to define features within an MILP solver. 
While current trends in optimization often rely on neural networks, these approaches have notable drawbacks, including their dependency on oracles, their opaque "black-box" nature, and their inconsistent performance, which may arise from the expansive search space and significant computational overhead. In contrast, methods like ours, with their simpler and more constrained search spaces, offer faster convergence to high-quality solutions. Additionally, our approach is designed to be highly efficient, imposing minimal additional cost on the overall solving process. Future work could explore the application of automatic heuristic search methods to other solver components, such as branching strategies or cutting plane selection.

\section*{Acknowledgements}  
This research was funded by the Agence Nationale de la Recherche (grant ANR-22-CE46-0011) and the Luxembourg National Research Fund (grant INTER/ANR/22/17133848) as part of the UltraBO Project. The computational experiments were conducted using the University of Luxembourg’s HPC facility.

We express our gratitude to Mark Turner for his support with SCIP and the MIPLIB 2017 library, and to Abdel Labassi for his guidance in using his framework. We are also deeply grateful to Pierre Talbot for the valuable informal discussions that greatly improved the quality of this work.  Finally, we appreciate the constructive feedback from the anonymous reviewers, which significantly contributed to this revised version. 

\appendix
\section*{Appendices}
\section{Hardware and Software Settings}\label{hardware_appendix}
We used SCIP 9.1 along with the DEAP 1.4.1 package as the framework for implementing our GP algorithm. Additionally, we integrated the method from~\cite{labassi_learning_2022} as a baseline. For their method, we retained all parameters used in the original paper. This baseline relies on PyTorch, and we used PyTorch 2.4.0 for our implementation.

To specify the hardware setup used in each simulation phase, we utilized the following:
\begin{itemize}
    \item For the first phase, requiring GPU resources: 1 Xeon Gold 6132 @ 2.6GHz [14c/140W], 2 Tesla V100 SXM2 16G.
    \item For the second phase, which only required CPU resources: either 2 Xeon E5-2680v4 @ 2.4GHz [14c/120W] or 2 Xeon Gold 6132 @ 2.6GHz [14c/140W] nodes.
\end{itemize}

\section{GP  Behavior for Specific Problem Types}~\label{GP_three_problems}
We applied GP methods to the FCMCNF, MAXSAT, and GISP problems. Due to the extensive runtime of each GP method, we did not conduct robustness tests by repeatedly running GP on the same problem. However, the SS policies derived for each problem type can be considered effective, as they demonstrate strong performance compared to the other baselines.

\begin{figure*}[htbp]
\footnotesize
\centering 
{%
\subfigure [] {
\includegraphics[scale=0.33]{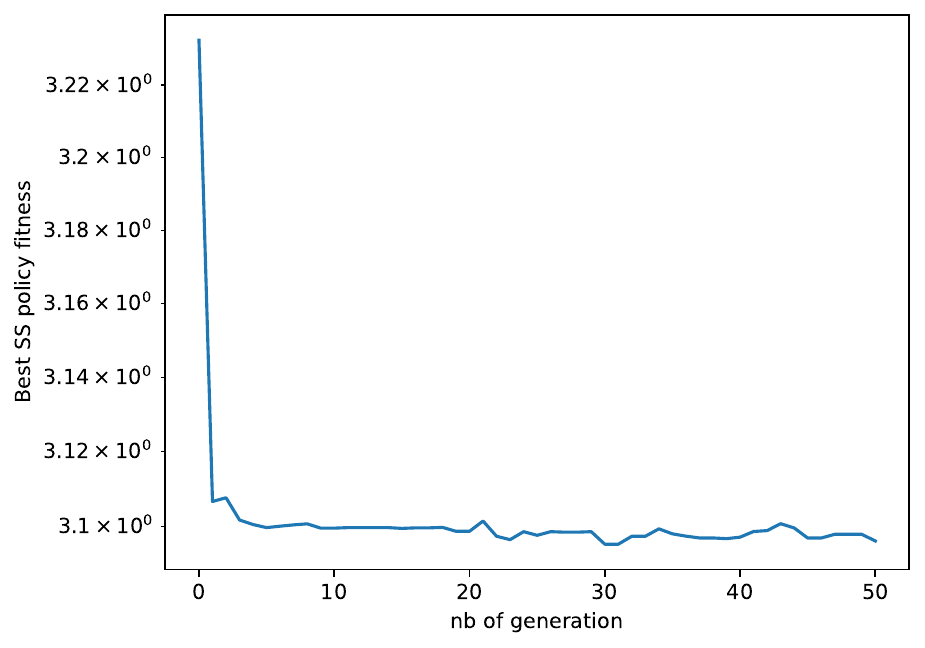}
    }
\hfill
\subfigure [] {%
    \includegraphics[scale=0.33]{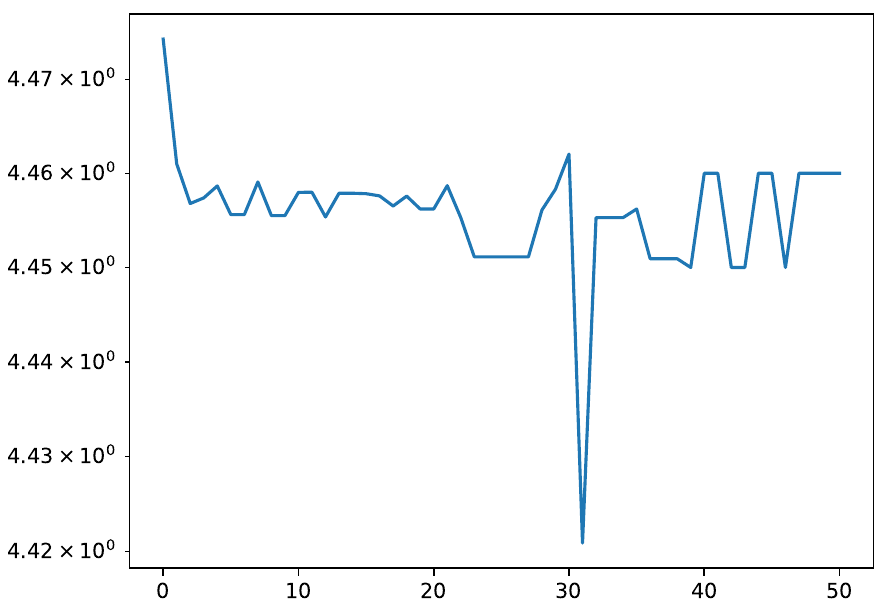}
    }
\hfill
\subfigure [] {
    }
\includegraphics[scale=0.33]{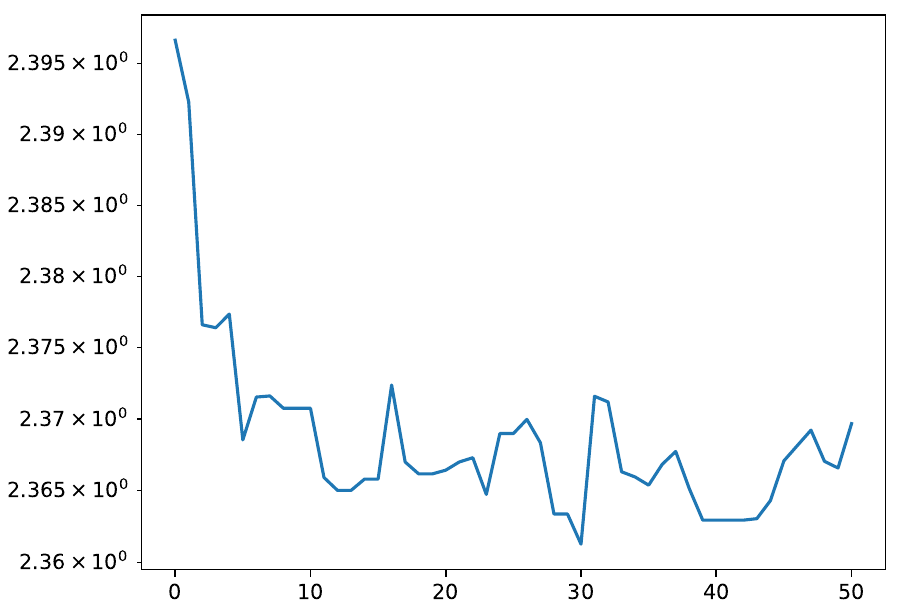}
    }
\caption{
1-shifted geometric mean solving time over the training set of the best SS policy in the population across generations for: (a) FCMCNF, (b) MAXSAT, and (c) GISP.}
\label{convergence_GP_three_problems}
\end{figure*}

In~\cref{convergence_GP_three_problems}, we present convergence graphs showing the fitness function (1-shifted geometric mean solving time) of the best individual among the 50 considered for each generation. These graphs allow us to assess the convergence speed of the method towards the optimal solution. Additionally, in~\cref{gp_func_three_problems}, we display the scoring function that performed best on the training set for each problem studied. Notably, as seen in~\cref{convergence_GP_three_problems}(a), the best scoring method for FCMCNF converges very quickly with little subsequent improvement. This result is due to the final method representing a best estimate BFS; the convergence is consistent, as seen in~\cref{simulation_artificial}, where this heuristic is the best-performing baseline. For the other methods, the final function is slightly more complex (using three terminals rather than two) and is found after a greater number of generations. Interestingly, for MAXSAT and GISP, the \emph{GP2S} method prioritizes higher values of the best estimate (since they are divisors in the formula).

\begin{table}[htbp]
    \centering
    \begin{tabular}{|l|l|}
     \hline
         Problem& GP function  \\\hline\hline
         FCMCNF& $\frac{BE_i}{d_i}$ \\\hline
         MAXSAT &  $\frac{m}{BE_i + M}$\\\hline
         GISP & $\frac{d_i}{BE_i}-d_i$\\\hline
    \end{tabular}
    \caption{Scoring functions produced by GP from training sets for three problem types.}
    \label{gp_func_three_problems}
\end{table}

\section{GP Scoring Functions for MIPLIB 2017}~\label{GP_MIPLIB}

\begin{table}[htbp]
    \centering
    \begin{tabular}{|l|m{5cm}|}
     \hline
 \textbf{Seed} & \textbf{GP Function}  \\\hline\hline
         $1$ & $z_i+ \frac{n}{d_i- n} $\\\hline
$2$ & $BE_i+ z_i+ m$\\\hline
$4$ & $\frac{z_i}{z_0}+ n\times d_i$\\\hline
$6$ & $(BE_i+ z_i)\times (BE_i+ z_i)$\\\hline
$8$ & $z_0\times(M+M- z_i)- BE_i- d_i\times z_i$\\\hline
$10$ & $\frac{z_i}{\frac{m}{d_i}}$\\\hline
$12$ & $\frac{z_i}{d_i\times ( n+ BE_i+ d_i)}$\\\hline
$14$ & $z_i \times\frac{\frac{d_i}{z_i}}{z_i}\times  (BE_i- n)$\\\hline
$16$ & $z_i+ z_i$\\\hline
$18$ & $M-z_i\times d_i+ BE_i$\\\hline
$20$ & $\frac{\frac{n- m}{M}}{BE_i}$\\\hline
    \end{tabular}
    \caption{Functions generated by GP from training sets based on MIPLIB 2017 instances.}
    \label{gp_MIPLIB}
\end{table}
We now present the functions generated by GP across the 10 seeds in~\cref{gp_MIPLIB}. It is important to note once again that the set of instances has a significant impact on the convergence of the GP method. We observe that, due to the increased heterogeneity of the instances (including varying instance sizes and problem types), the final scoring functions are considerably more complex, incorporating a greater number of problem-related variables.

\bibliography{references}

\end{document}